# Incorporating Machine Learning to Evaluate Solutions to the University Course Timetabling Problem

**Patrick Kenekayoro**

Niger Delta University, Amassoma, Nigeria
Patrick.Kenekayoro@outlook.com

*Abstract*—Evaluating solutions to optimization problems is arguably the most important step for heuristic algorithms, as it is used to guide the algorithms towards the optimal solution in the solution search space. Research has shown evaluation functions to some optimization problems to be impractical to compute and have thus found surrogate less expensive evaluation functions to those problems. This study investigates the extent to which supervised learning algorithms can be used to find approximations to evaluation functions for the university course timetabling problem. Up to 97 percent of the time, the traditional evaluation function agreed with the supervised learning regression model on the result of comparison of the quality of pair of solutions to the university course timetabling problem, suggesting that supervised learning regression models can be suitable alternatives for optimization problems' evaluation functions.
*Keywords/Index Terms*—classification, genetic algorithm, incremental learning, machine learning, university course timetabling problem

## 1. Introduction

Evaluating solutions to optimization problems is one of the most frequent and arguably the most important operation in meta-heuristic algorithms, as evaluation is necessary to guide meta-heuristic algorithms to optimal solutions in the solution search space. If the evaluation function for a certain optimization problem is computationally expensive, the number of evaluations per unit time is reduced, thereby limiting the extent to which the solution search spaced is traversed to find good solutions. Thus, there is a need to explore alternate less expensive methods for evaluating the quality of solutions to optimization problems that can be used in lieu of the traditional more expensive evaluation function. Booker et al. (1999) have presented a framework for identifying surrogate inexpensive approximations of a more expensive evaluation function.

Classification and regression are supervised learning techniques that have been used in a number of research fields (Kenekayoro et al., 2014; Kotsiantis et al., 2004) for predictions based on previously learnt information, and so, it will be interesting to investigate the extent to which supervised learning can be used to identify approximations to heuristics evaluation functions. Thus, the main focus of this research is to demonstrate the possibility of using machine learning techniques for the evaluation of optimization problems.

Timetabling problems are a class of well-known optimization problems that have been proven to be NP Complete in 5 different ways (Cooper & Kingston, 1996). The University Course Timetabling Problem (UCTP) is one of such NP Complete optimization problems faced annually in all Higher Education Institutions. Techniques to solving this problem have been researched extensively, some of which include meta-heuristic techniques such as simulated annealing, hill climbing, tabu search and evolutionary methods such as particle swarm optimization, ants or bee colony optimization techniques (Abayomi-Alli et al., 2019; Abdullah et al., 2010; Bolaji et al., 2011; Hao & Benlic, 2011; Kalender et al., 2012; Kenekayoro, 2012; Socha et al., 2003; Song et al., 2018). Adewumi et al. (2016) have further designed a mobile based system that improves the presentation for constructed timetables for end users. The extensive research on the UCTP makes it suitable for investigating the extent to which machine learning methods can be used to evaluate solutions to optimization problems.

The main aim of this research is to investigate the extent to which machine learning methods may be used to create an alternate evaluation function for computational expensive evaluation functions. To this end;

- The genetic algorithm is used to generate solutions and their corresponding fitness (by an evaluation function) to the University Course Timetabling Problem, and
- Supervised machine learning algorithms are then used to create learning models to determine if these models can accurately approximate the evaluation function.

Subsequent sections give an overview of the UCTP, describes how the genetic algorithm was used to generate examples for the supervised learning algorithms, and then results of training the supervised learning algorithms are presented.

## 2. Literature Survey
### 2.1. University Course Timetabling Problem

The University Course Timetabling Problem is concerned with creating a schedule containing events that meet several constraints. The constraints could be hard or soft. Hard constraints are those conditions that must be met, while soft constraints are the desired conditions that may not be met. A violation of a hard constraint results in an infeasible schedule. The International Timetabling Competition in 2007 (ITC-2007) elaborately defined the generic constraints for the UCTP and made available datasets from real universities that can be used to test algorithms. This dataset has been used to determine the performance of recent heuristic algorithms for solving the UCTP (Akkan & Gülcü, 2018; Nagata, 2018). The following constraints for the ITC-2007 as described by Bonutti et al. (2012) are used in this study:

**Hard Constraints**
HC1. All weekly lectures of a course must be assigned to a distinct period.
HC2. A room cannot be assigned to more than one lecture in a given period.
HC3. Courses belonging to a curriculum must be assigned to different periods.
HC4. Courses taught by a lecturer must be assigned to different periods.
HC5. Courses taught by a lecturer must not be assigned to periods when the lecturer is unavailable.

**Soft Constraints**
SC1. The number of students taking a course must fit into the assigned room. Each additional student over the capacity of the room counts as a violation.
SC2. Lectures of a course must be spread between the minimum days for that course. Each day less than the minimum days has a penalty of 5.
SC3. Lectures for a given curriculum must be in consecutive.
SC4. All lectures of a given course must be in the same room. Each additional room allocated to a lecture of a given course counts as a violation.

There are a number of techniques that have been used to find good solutions to timetabling problems (Babaei et al., 2014). Particularly, the adaptive tabu search (Lü et al., 2010) has found the best known solutions to a number of the instances in the ITC-2007 dataset. However, as this research is concerned with finding an alternative less computationally expensive evaluation function for solutions to optimization problems, the Genetic Algorithm (Bäck et al., 2000) is appropriate for creating a dataset that can be used for this investigation. An individual in a genetic algorithm represents a solution to the problem the genetic algorithm solves. The genetic algorithm has also been successfully used to solve timetabling problems (Bhatt, 2004; Lewis & Paechter, 2007; Weare et al., 1995), although the GA may not always be the most suitable algorithm if the main goal is finding the most optimal timetable solution (Lewis & Paechter, 2007).

## 2.2. Genetic Algorithm
The genetic algorithm is modelled on natural evolution. An initial population made up of individuals (solutions to an optimization problem) form the first generation, and then over the course of evolution, subsequent generations are determined by genetic operators (selection, mutation and crossover). The assumption is that as the population evolves, the individuals become fitter; that is better solutions are found. Algorithm 1 shows a simplified pseudo code for the genetic algorithm.

**ALGORITHM 1** PSEUDO CODE FOR THE GENETIC ALGORITHM

```
1. Generate initial population
2. Save fittest individual
3. While stopping condition not met
   4. Create offsprings from parents by
      mutation and/or crossover genetic
      operators
   5. Select individuals from parents and
      offsprings to form the population for
      next generation
   6. Update fittest individual
7. Return fittest individual
```

An approach to solving constraint satisfaction problems like the UCTP that have required and desired constraints is separating the algorithm into two stages, the first for finding a feasible solution, while the second for reducing the number of soft constraint violations found in the feasible solution (Yang & Jat, 2011). This technique is particularly useful for the UCTP because feasible solutions can often be found by low level heuristics such as least saturation degree first (Abdullah et al., 2010; Bolaji et al., 2011; Tuga et al., 2007) or other ordering heuristics (Burke et al., 2007). The genetic algorithm is flexible in that the hard and soft constraint evolution stages can use identical genetic operators and/or evolution strategies with difference only in the representation of individuals and how they are transformed to a human readable timetable solution. This study uses two stages to find optimal solutions to the UCTP problems in the ITC-2007 dataset, but because the difference in representation and genetic operators is minimal both representations and genetic operators for hard and soft constraint evolution are described in the same section.

**Representation**

A solution to a university course timetabling problem (individual) is represented as a list of integers. The length of the individual is determined by equation 1, which is the number of lectures necessary to meet the first hard constraint (HC1). For individuals in the hard constraint evolution, the possible allele values are integers in the range [0, number of lectures), while possible allele values for individuals in soft constraint evolution are integers in the range [0, number of room period pairs). The number of room period pairs is determined by equation 2.

The representation of individuals is the main difference between the genetic algorithm used to find optimal solutions for soft and hard constraints. The representations also determine how the human readable timetable solutions are generated.

In the hard constraint representation, allele values represent the order in which events are scheduled. To create a timetable solution for an individual represented as [2, 1, 3, 0] (hard constraint representation) for the timetabling problem in Figure 1 and event list in Table 1, event C21 is assigned to its best fitting room period pair, and then C12 is assigned to its best fitting room period pair that is not already taken, followed by C22, and lastly C11.

In the soft constraint representation, the position of the allele in the individual represents the index of the event in the lecture list, while the value of the allele represents the index of the room period pair in the room period list. An example individual for the timetabling problem, lecture list and room period pairs from Figure 1, Table 1 and Table 2 is shown in Table 3.

$$NumEvents \sum_{i=1}^{numcourses} numLecturesInWeek(Course_i) \quad (1)$$

$$NumRoomPeriodPairs = NumRooms * DaysInWeek * PeriodsInDay \quad (2)$$

**Neighborhood Functions**

Neighborhood functions are low level operators that transform a solution to an optimization problem to a new solution. The chosen higher level algorithm determines how neighborhood functions are used to guide the algorithm to an optimal solution. This research uses two low level operators.

| COURSE | NO. LECTURES IN WEEK |
|---|---|
| C1 | 2 |
| C2 | 2 |
| ROOMS | |
| R1 | R2 |
| NO. DAYS IN WEEK | NO. PERIODS IN DAY |
| 2 | 2 |

Figure 1. A toy example of possible courses, rooms and periods in a university timetabling problem

Table 1 List Of Events For Courses In The University Course Timetabling Problem In Fig 1 And Their Indices In A List

| $C1_1$ | $C1_2$ | $C2_1$ | $C2_2$ |
|---|---|---|---|
| 0 | 1 | 2 | 3 |

Table 1 List of room period combinations for the course timetabling problem in Figure 1 and their indices in a list

| $R1_0$ | $R1_1$ | $R1_2$ | $R1_3$ | $R2_0$ | $R2_1$ | $R2_2$ | $R2_3$ |
|---|---|---|---|---|---|---|---|
| 0 | 1 | 2 | 3 | 4 | 5 | 6 | 7 |

Table 2 Generated schedule for an individual's soft constraint representation [7, 0, 3, 1] for the timetabling problem in Figure 1

| Course | Room | Day | Period in Day | Period in week |
|--------|------|-----|---------------|----------------|
| C1 | R2 | 1 | 1 | 3 |
| C1 | R1 | 0 | 0 | 0 |
| C2 | R1 | 1 | 1 | 3 |
| C2 | R1 | 0 | 1 | 1 |

**Simple move**

Sets the gene value in the $i^{th}$ position of an individual to a new value. If the new value is already taken, swap the gene value of that position with the gene value of the $i^{th}$ position.

Algorithm 2 Pseudo Code for Simple Move

```
1    simple_move (individual, position_one, new_allele)
2        If new_allele in individual
3            position_two = index of new allele in individual
4            swap gene values at position_one and position_two
5        else
6            set gene value at position_one to new_allele
```

**Chain move**

A chain move is a variant of the kempe chain interchange (Morgenstern & Shapiro, 1990); a well-known neighborhood function used in a number of studies that investigated heuristics techniques to solve timetabling problems (Chiarandini et al., 2006; Lewis, 2006; Lü & Hao, 2008). Lu and his colleagues (2010) have succinctly described it as moving an event to a new period and then moving additional events that cause a violation as a result of the previous move. In this study, chain moves are only used in soft constraint evolution.

Algorithm 3 Pseudo Code for Chain Move

```
1    chain_move ( individual, position, new_allele)
2        event_one = event in index position of event list
3        p1 = period of room period pair in index new_allele of room period list
4        if moving event to room period pair in room_period_list[new_allele] causes a room, student or lecturer clash
5            p2 = get old period of event_one
6            move_list_one = get events clashing with event_one in p2
7            move_list_two = get events clashing with the events in move_list_one in p1
8            move events in move_list_one to period
```

```
                    p2
9                   move events in move_list_two to period
                    p1
10              else
11                  set gene value at position to
                    new_allele
```

## Genetic Operators
It is agreed that genetic operators greatly affect the performance of genetic algorithms (Murata & Ishibuchi, 1996), and studies (Srinivas & Patnaik, 1994) have investigated optimal ways to choose the probability of applying a crossover or mutation operator to create a new individual. Evolution strategies such as the basic genetic algorithm (Bäck et al., 2000) and the covariance matrix adaptation (Hansen & Ostermeier, 2001) describe strategies that can be used to create a new generation from the current using mutation and crossover operators.

## Mutation
Mutation changes the gene value of one or more alleles in an individual. The mutation operator used in this study's genetic algorithm that solves the timetabling problem described in the ITC-2007 changes a random gene value to a new possible value. If that gene value is already taken, then both alleles are swapped. This is implemented through the simple move and chain move neighborhood functions. The hard constraint mutation only makes use of the simple move operator while the soft constraint mutation alternates between the simple move and the chain move functions after a number of consecutive non improving generations.

## Crossover
Crossover combines two individuals in a population to create two offsprings. Single point, two point and uniform are some kind of crossover operators. Magalhaes-Mendes (2013) elaborately described these crossover operators and the results of the experiments Magalhaes-Mendes (2013) carried out suggests that the single point crossover performs better than other crossover operators for the job shop scheduling problem. This may not necessarily be true for the university course timetabling problem in this study, but the single point crossover is used in this study.

In a single point crossover of two individuals, the first n/2 alleles of the first individual is replaced with the second individual's first n/2 operators to create the first offspring. To create the second offspring, the second n/2 alleles of the first individual is replaced with the second n/2 alleles of the second individual, where n is the length of the individual.

| 1 | 2 | 3 | 4 | 5 | 6 | Parent One |
| 7 | 8 | 9 | 10 | 11 | 12 | Parent Two |
|  |  |  |  |  |  |  |
| 7 | 8 | 9 | 4 | 5 | 6 | Offspring One |
| 1 | 2 | 3 | 10 | 11 | 12 | Offspring Two |

Figure 2 An Example Of A Single Point Crossover Operator

Algorithm 4 shows the pseudo code that implements the single point crossover operator with neighborhood functions that is used in this research. As with the mutation operator, chain moves and simple moves are alternated after consecutive non improving generations in soft constraint evolution, while hard constraint evolution utilizes only the simple move neighborhood function.

**Selection**

The selection operator determines the individuals that will make up the population that of the next generation. For a specified number of tournaments, tournament selection selects a random set of k individuals from the parents and offsprings and then the best individual in the tournament is added to the individuals that make up the subsequent generation.

The roulette wheel selection strategy is modelled on the casino roulette wheel. Individuals that make up the subsequent generation are selected based on a probability proportional to their fitness. The probability of a fitter individual being selected is higher than that of a less fit individual.

A number of variations to the roulette wheel and tournament selection strategies exist and there is no consensus at to which selection strategy is the most appropriate. For example, in the travelling salesman problem, tournament selection is suitable for smaller input sizes, while ranked roulette wheel is appropriate for larger sized problems (Razali & Geraghty, 2011). Thus, a researcher has to decide on an appropriate selection strategy on a case by cases basis. Tournament selection is used in this study.

ALGORITHM 2 PSEUDO CODE FOR SINGLE POINT CROSSOVER

```
1    single_point_crossover(ind_one, ind_two, chain)
2        half_point = length of individual / 2
3        for index in range(0,  half_point)
4           if chain
5              offspring_one = chain_move(ind_one, index, ind_two[index])
6           else
7              offspring_one = simple_move(ind_one, index, ind_two[index])
8        for index in range(half_point, length of individual)
9           if chain
10             offspring_two = chain_move(ind_two, index, ind_one[index])
11          else
12             offspring_two = simple_move(ind_two, index, ind_one[index])
```

**Evaluation**

The quality of an individual is determined by the number of hard constraint violations listed in the earlier sections. Lu, Hao and Zhipeng (2010) have elaborately defined these constraints mathematically.

During the soft constraint evolution, the weight of each hard constraint violation is increased by 1000 in order to guide the algorithm towards only feasible solutions.

## GA Strategy

The Distributed Evolutionary Algorithms in Python – DEAP (Fortin et al., 2012) includes an implementation of the genetic algorithm as well as other evolutionary algorithms. The DEAP python package is flexible in that it provides the building blocks to create custom genetic operators or evolution strategies, which is why it is used in this research. The genetic algorithm in this study evolves the population until the quality of the solution is better than those published in (Bolaji et al., 2011) and 100 consecutive non improving generations reached. The evolution strategy uses tournament selection with the varOr algorithm which generates a specified number of offsprings by mutation, crossover or reproduction by specified probability (Fortin et al., 2012).

Even though this research solves the UCTP, it is important to stress that the main aim of this study is not solving the UCTP but identifying an alternative method to efficiently evaluate solutions to constraint satisfaction problems that may otherwise be computationally expensive to evaluate. The dataset that will be used to investigate how well regression and classification supervised learning techniques can be used as alternative evaluation function for constraint satisfaction problems is generated over the course of the evolution of the genetic algorithm.

## 2.3. Supervised Learning

Supervised learning uses patterns identified in a training dataset to map variables or features to labels. When the labels are categorical variables the supervised learning task is said to be classification, while the task is said to be regression if the labels are continuous variables.

The individuals created over the course of evolution with the genetic algorithm can form the dataset for machine learning evaluation. Gene values (alleles) are the variables the supervised learning algorithm can use to map individuals to its fitness; a regression task. Determining if an individual is a feasible (F) solution or non-feasible (NF) solution is a two class classification task.

Incremental learning is a method that updates a supervised learning prediction model with new training examples without losing information learnt from past training examples. Polikar et al. (2001) listed that an incremental learning algorithm

*"must be able to learn new information from new data, must not require access to old data to update the prediction model, must preserve previously learnt patterns and must be able to accommodate new classes that may be introduced with new data"*

Intuitively, incremental learning seems appropriate for creating predictive models that can be used as an alternative evaluation function during a genetic algorithm's evolution because new data is introduced as new individuals are created during evolution, hence it makes sense to update the predictive model with information from the newly created individuals. However, it can also be argued that individuals evolve with each generation as the next generation is usually fitter than the previous generation. Past information is not necessarily needed to correctly predict future examples. Incremental learning may even affect the accuracy of the model, if the classification algorithm is too stable, that is the algorithm preserves information from past data while not learning enough new information from new data (Polikar et al., 2001).

Supervised learning algorithms for regression and classification are implemented in the scikit-learn machine learning python package (Pedregosa et al., 2011). The support vector machines (Cortes & Vapnik, 1995) is an appropriate supervised learning algorithm that can be used for regression and classification as it has been shown to be one of the best non-ensemble supervised learning algorithms (Caruana & Niculescu-Mizil, 2006). Also, support vector machines is among the algorithms implemented in scikit-learn that can be used for incremental learning. In this study, support vector machines is used for classification, regression and incremental learning.

## 3. Methods

The genetic algorithm is used to generate a dataset which is then used to train supervised learning methods to approximate the evaluation function of the UCTP.

### 3.1 Dataset Generation

To generate the training and test dataset for supervised learning evaluation, the genetic algorithm is used to solve the UCTP problem for the ITC-2007 dataset. On each evaluation of an individual, the individual, its fitness value and the category it belongs to (feasible or non-feasible) is saved as an instance for training the machine learning algorithms. The category label is excluded for the regression task, while the fitness value is excluded for the classification task.

Table 4 Description Of The Ga Generated Dataset For The 21 Itc-2007 Problem Instances

| Dataset | No. Instances | Min | F (%) | NF (%) |
|---------|---------------|-----|-------|--------|
| Comp01 | 466838 | 31 | 62.5 | 37.5 |
| Comp02 | 619383 | 306 | 73.2 | 26.8 |
| Comp03 | 512257 | 320 | 71.8 | 28.2 |
| Comp04 | 668704 | 192 | 66.9 | 33.1 |
| Comp05 | 732832 | 651 | 65.5 | 34.5 |
| Comp06 | 737495 | 304 | 76.7 | 23.3 |
| Comp07 | 930689 | 314 | 68.9 | 31.1 |
| Comp08 | 615967 | 223 | 69.5 | 30.5 |
| Comp09 | 639525 | 246 | 66.5 | 33.5 |
| Comp10 | 698410 | 273 | 74.8 | 25.2 |
| Comp11 | 427531 | 25 | 67.2 | 32.8 |
| Comp12 | 568963 | 649 | 63.4 | 36.6 |
| Comp13 | 407604 | 320 | 82.4 | 17.6 |
| Comp14 | 589972 | 226 | 70.0 | 30.0 |
| Comp15 | 669928 | 273 | 72.3 | 27.7 |
| Comp16 | 844754 | 274 | 81.5 | 18.5 |

| Comp17 | 575094 | 295 | 77.6 | 22.4 |
| --- | --- | --- | --- | --- |
| Comp18 | 571946 | 169 | 59.6 | 40.4 |
| Comp19 | 607892 | 283 | 77.1 | 22.9 |
| Comp20 | 767332 | 445 | 75.7 | 24.3 |
| Comp21 | 851448 | 327 | 72.6 | 27.4 |

The performance of the genetic algorithm for finding good solutions to the university course timetabling problems in the ITC-2007 dataset is shown in the min column of Table 4. The results achieved for the 21 instances in the ITC-2007 dataset is not among the best published, but they are competitive in the sense that the genetic algorithm can still find better solutions if the evolution continues, and results are better than the solutions by some previous algorithms (Bolaji et al., 2011; Wahid, 2014). The genetic algorithm has also generated enough samples that can be used to investigate how well regression and classification can be used for evaluating the quality solutions, which is the main goal if this research.

## 4. Results and Discussion

The result for regression shows how well the supervised learning algorithms in this study approximated the UCTP evaluation function, while classification showed the extent to which the supervised learning algorithm can determine if a solution is feasible or infeasible.

### 4.1 Regression

In the majority of heuristic algorithms, the quality of solutions are compared to guide the algorithm to the optimal solution in the search space. So, even though mean absolute error and its variants are amongst the most commonly used evaluation metrics for supervised learning regression models (Baccianella et al., 2009), the quality of the regression model in this study is determined by evaluating the extent to which comparison with the regression model is the same as the comparison with the UCTP evaluation function.

The dataset is split into two disjoints sets, 70% for training and 30% for tests. Each example in the test set is compared with all other examples using the UCTP evaluation function and the trained regression model. Thus, if the training set contains 3 examples there are 3 combination 2 $\{E_{0,1}, E_{0,2}, E_{1,2}\}$ comparisons. The value of the comparisons is determined by Equation 3. The overall accuracy of regression model is determined by Equation 4.

$$Compare(E_{i,j}) = \begin{cases} 1, & Evaulate(E_i) \leq Evaluate(E_j) \\ 0, & Evaluate(E_i) > Evaluate(E_j) \end{cases} \quad (3)$$

$$accuracy = \frac{\sum_{1}^{n} Compare_r(x_{i,j}) = Compare_e(x_{i,j})}{n} \quad (4)$$

$n = total\ number\ of\ comparisons$
$r = comparison\ by\ regression\ model$

$e = comparioson\ by\ evaluation\ model$

Table 5 shows the accuracy of the regression model for the dataset in Table 4, generated by the genetic algorithm. All 21 datasets contained more 400,000 example instances as shown in Table 4, thus the data is split into batches of 10,000 instances for training. This to some extent mimics the real life scenario when new data is generated as a heuristic algorithm progresses, albeit with a smaller batch size. At the first glance, incremental learning seems to be the appropriate technique to train the supervised learning model because the training is done in batches. However on closer inspection, and based on the results shown in Table 5 traditional learning performs better than incremental learning because as the algorithm progresses the data also evolves and so past information learnt may sometimes limit the learning of new information found in a newly introduced batch.

TABLE 3 ACCURACY OF REGRESSION MODEL FOR PREDICTION WITH TRADITIONAL LEARNING AND INCREMENTAL LEARNING

|         | Incremental learning | | | Traditional Learning | | |
| --- | --- | --- | --- | --- | --- | --- |
| **Dataset** | **Min** | **Max** | **Mean** | **Min** | **Max** | **Mean** |
| comp01 | 0.6 | 0.95 | 0.82 | 0.62 | 0.98 | 0.93 |
| comp02 | 0.79 | 0.96 | 0.77 | 0.79 | 0.99 | 0.94 |
| comp03 | 0.51 | 0.96 | 0.79 | 0.83 | 0.99 | 0.95 |
| comp04 | 0.57 | 0.96 | 0.81 | 0.73 | 0.99 | 0.94 |
| comp05 | 0.51 | 0.94 | 0.77 | 0.83 | 0.99 | 0.95 |
| comp06 | 0.51 | 0.93 | 0.75 | 0.83 | 0.99 | 0.96 |
| comp07 | 0.42 | 0.88 | 0.63 | 0.82 | 0.99 | 0.93 |
| comp08 | 0.65 | 0.94 | 0.8 | 0.83 | 0.99 | 0.94 |
| comp09 | 0.56 | 0.97 | 0.81 | 0.78 | 0.99 | 0.94 |
| comp10 | 0.57 | 0.93 | 0.77 | 0.43 | 0.99 | 0.94 |
| comp11 | 0.57 | 0.97 | 0.81 | 0.52 | 0.99 | 0.92 |
| comp12 | 0.57 | 0.92 | 0.79 | 0.82 | 0.99 | 0.94 |
| comp13 | 0.63 | 0.95 | 0.81 | 0.92 | 0.99 | 0.97 |
| comp14 | 0.55 | 0.96 | 0.8 | 0.75 | 0.99 | 0.95 |
| comp15 | 0.51 | 0.96 | 0.78 | 0.62 | 0.99 | 0.95 |
| comp16 | 0.52 | 0.94 | 0.74 | 0.94 | 0.99 | 0.97 |
| comp17 | 0.59 | 0.96 | 0.78 | 0.68 | 0.99 | 0.94 |
| comp18 | 0.58 | 0.94 | 0.81 | 0.44 | 0.99 | 0.94 |
| comp19 | 0.53 | 0.96 | 0.79 | 0.73 | 0.99 | 0.96 |
| comp20 | 0.51 | 0.83 | 0.69 | 0.82 | 0.99 | 0.96 |
| comp21 | 0.53 | 0.97 | 0.77 | 0.58 | 0.99 | 0.94 |

Up to 97% average accuracy was obtained for the traditional regression model. This high accuracy suggests that to some extent, the model can be used in lieu of the standard evaluation function, particularly for algorithms such as the hill climbing or simulated annealing whose traversal of the solution search space is based on comparisons of neighborhood solutions.

For well-known problems with computationally expensive evaluation functions, a pre-trained regression model can be made publicly available along with the dataset so that researchers can speedily test algorithm prototypes with this pre-trained model, without the bottleneck of an expensive evaluation function.

In as much as the regression model can be used to accurately predict the result of comparing solutions to the optimization problems, it cannot tell if a solution is feasible or infeasible. Moreover, for the problems in the ITC-2007 dataset, an infeasible solution can still have a smaller number of violations. Predicting if a solution is feasible is a two class classification problem.

### 4.2. Classification

Three metrics simple accuracy, recall and precision are used to evaluate the quality of the classification models. While simple accuracy is simply the ratio of correctly classified instances to the total number of instances, precision shows how well a classification model correctly predicts individual classes and recall shows how well the classification model does not misclassify instances.

TABLE 4 AVERAGE ACCURACY (A), PRECISION (P) AND RECALL (R) FOR PREDICTING THE FEASIBILITY OF SOLUTIONS TO UNIVERSITY COURSE TIMETABLING PROBLEMS USING TRADITIONAL LEARNING WITH SUPPORT VECTOR MACHINES AND INCREMENTAL LEARNING WITH STOCHASTIC GRADIENT DESCENT

|  | Incremental Learning | | | Traditional Learning | | |
|---|---|---|---|---|---|---|
| Dataset | A | P | R | A | P | R |
| comp01 | 0.79 | 0.79 | 0.79 | 0.94 | 0.94 | 0.94 |
| comp02 | 0.77 | 0.72 | 0.77 | 0.95 | 0.95 | 0.95 |
| comp03 | 0.80 | 0.78 | 0.80 | 0.95 | 0.95 | 0.95 |
| comp04 | 0.76 | 0.74 | 0.76 | 0.93 | 0.93 | 0.93 |
| comp05 | 0.74 | 0.71 | 0.74 | 0.95 | 0.95 | 0.95 |
| comp06 | 0.72 | 0.68 | 0.72 | 0.95 | 0.95 | 0.95 |
| comp07 | 0.77 | 0.72 | 0.77 | 0.95 | 0.95 | 0.95 |
| comp08 | 0.79 | 0.76 | 0.79 | 0.93 | 0.94 | 0.93 |
| comp09 | 0.80 | 0.77 | 0.80 | 0.94 | 0.94 | 0.94 |
| comp10 | 0.75 | 0.71 | 0.75 | 0.95 | 0.96 | 0.95 |
| comp11 | 0.81 | 0.79 | 0.81 | 0.92 | 0.92 | 0.92 |
| comp12 | 0.79 | 0.78 | 0.79 | 0.96 | 0.95 | 0.96 |
| comp13 | 0.75 | 0.71 | 0.75 | 0.94 | 0.94 | 0.94 |
| comp14 | 0.76 | 0.72 | 0.76 | 0.95 | 0.95 | 0.95 |
| comp15 | 0.81 | 0.80 | 0.81 | 0.95 | 0.95 | 0.95 |
| comp16 | 0.72 | 0.67 | 0.72 | 0.95 | 0.95 | 0.95 |
| comp17 | 0.76 | 0.70 | 0.76 | 0.95 | 0.95 | 0.95 |
| comp18 | 0.87 | 0.86 | 0.87 | 0.93 | 0.93 | 0.93 |
| comp19 | 0.72 | 0.69 | 0.72 | 0.95 | 0.95 | 0.95 |
| comp20 | 0.70 | 0.66 | 0.70 | 0.95 | 0.96 | 0.95 |
| comp21 | 0.79 | 0.75 | 0.79 | 0.95 | 0.95 | 0.95 |

## 5. Conclusions

The evaluation function for some optimization problems can sometimes be computationally impractical to evaluate, and so studies have researched ways to identify surrogate approximations of these computationally expensive functions. This study has described in detail the procedure for solving the university course timetabling problem with the genetic algorithm and has also demonstrated that supervised learning techniques are suitable for finding alternate less expensive functions for a more computationally expensive evaluation function that can be used in heuristic algorithms such as simulated annealing or the genetic algorithm.

The regression model in this study obtained up to 97% agreement with the traditional evaluation function for comparing the quality of pair of solutions to the university course timetabling problem. On its own, a regression model is not sufficient to determine if a solution is feasible or infeasible, thus it is necessary to use supervised learning classification to predict if a solution is feasible. The support vector machines classification model could correctly predict if an instance is feasible or infeasible with up to 95% accuracy.

New solutions to optimization problems are generated as the heuristic algorithm progresses, and so incremental learning seems to be the natural technique to continuously update a prediction or regression model with the new solutions as they are generated. Surprisingly, traditional learning that ignored previous information outperformed incremental learning when training was in batches. This perhaps is because a new batch may not need past information to achieve generalization.

The methods used in this study can be used to create pre-trained models which can be incrementally updated as approximations for complex evaluation functions. This will enable researchers to more speedily test algorithm prototypes that solve optimization problems.